# A Generalized Approach for Cancellable Template and Its Realization for Minutia Cylinder-Code

Xing-Bo Dong, Zhe Jin and KokSheik Wong
School of Information Technology, Monash University Malaysia
E-mail: {xingbo.dong, jin.zhe, wong.koksheik}@monash.edu    Tel: +60-3 55145813

*Abstract*—Hashing technology has gained much attention in protecting the biometric template lately. For instance, Index-of-Max (IoM), a recently reported hashing technique, is a ranking-based locality sensitive hashing technique, which illustrates the feasibility to protect the ordered and fixed-length biometric template. However, biometric templates are not always in the form of ordered and fixed-length. Rather, it may be an unordered and variable size point set, e.g., fingerprint minutiae, which restricts the usage of the traditional hashing technology. In this paper, we propose a generalized version of IoM hashing, namely gIoM, to enable the utilization of unordered and variable size biometric template. We demonstrate a realization by using a well-known variable size feature vector - fingerprint Minutia Cylinder-Code (MCC). The gIoM transforms MCC into index domain to form indexing-based feature representation. Consequently, the inversion of MCC from the transformed representation is computational infeasible, thus achieving non-invertibility while the performance is preserved. Public fingerprint databases FVC2002 and FVC2004 are employed for experiments. Furthermore, the security and privacy analysis suggest that gIoM meets the criteria of template protection, namely, non-invertibility, revocability, and non-linkability.

## I. INTRODUCTION

Biometric Template Protection (BTP) is one of the crucial features in security and privacy protection for various biometric authentication and identification systems. Unlike password or tokens, biometric data strongly links with identity. When the biometric data is compromised, the leakage of the original biometric data is inevitable, which leads to privacy invasion. Therefore, biometric-based recognition systems with template protection are highly desirable. The idea of biometric template protection is to transform an unprotected template into a protected template so that attackers cannot recover the raw biometric data from the protected template. In the event of template compromise, a new template can be re-generated from the identical biometric source to replace the compromised one. To address this problem, many techniques have been reported. Generally, those techniques can be classified into two groups: *feature transformation* (or cancellable biometrics) and *biometric cryptosystem* [1]. Based on the characteristics of the transformation function used in feature transformation, this scheme can be further sub-categorized as salting and non-invertible transforms. Salting scheme refers that it concatenates a random key *r* with a secret key *k*, then stores the hash $H(r+k)$ in the database, where $H(\cdot)$ is a hashing function. In biometric salting, a user-specific and independent key such as password or random numbers is integrated with biometric data to generate a warped biometric template. On the other hand, one-way function is employed in non-invertible transformation schemes, thus it is computationally infeasible to retrieve original data even if any parameter of the transformation is revealed. In biometric cryptosystem scheme, there are mainly two approaches, namely, (a) generating a key from the biometric feature data (key generation), or (b) securing a key using biometric feature (key binding).

Among biometric traits, fingerprint is one of the most widely used and studied traits due to its long history of development, convenience in use and life-long sustainable patterns [2]. Although many fingerprint template protection schemes based on feature transformation have been proposed, it is still unsatisfactory to achieve a balance between performance and security. To design a good template protection scheme, the following criteria should be met:

• *Non-invertibility or Irreversibility:* It should be infeasible to retrieve the original biometric data, such as fingerprint minutiae points, from one single biometric template or multiple biometric templates. When the biometric template is compromised or stolen, the attacker cannot reconstruct the original feature based on the leaked template.

• *Revocability or Renewability:* It should be easy to issue a new protected template to replace the compromised one, and this also means it should be feasible to generate a very large number of protected templates from one original biometric template.

• *Non-linkability or Unlinkability:* It should be impossible to infer any information by matching two protected templates from two different applications. When multiple biometric databases are compromised, it should be infeasible for attackers to do the cross-match between different databases and retrieve the links between same individual.

• *Performance preservation.* The employing of template protection techniques should preserve the matching accuracy performance when compared to that of before-transformed counterparts.

The rest of the paper is organized as follows. First, literature review is carried out in Section II. In Section III, the motivations and contributions of the paper are highlighted. In Section IV, the relevant background knowledge is presented. Our proposed gIoM hashing scheme is described in Section V. Next, the experimental results are given in Section VI, supported by the performance analysis. In Section VII, security, privacy and revocability analysis are conducted. Finally, Section VIII concludes this paper.

## II. LITERATURE REVIEW

Over the last two decades, different template protection schemes have been proposed, and some review and survey papers already cover a comprehensive overview on this topic [1], [3], [4]. In this paper we focus on non-invertible feature transformation. A classification in the category of non-invertible feature transformation is proposed, namely Generic Hashing (GH) and Modality-Dependent hashing (MDH) according to the biometric modality to which the transformation can apply. GH is an independent biometric modality which can be apply to all biometric traits, including fingerprint, iris and face. On the other hand, MDH is dependent on the biometric modality, which can only be applied to certain biometric traits.

### A. Generic Hashing

Random Projection (RP) is a well-known generic transformation [5], [6]. RP is a process of projecting feature vector from $n$ dimension to $m$ ($n \gg m$) dimension in Euclidean space by random matrices. The theory of distance preservation for RP is proven by Johnson-Lindenstrauss lemma (J-L lemma) [7].

*J-L lemma:* Given any positive integer $p$, a positive number $k > 4\ln(p)/(\frac{\varepsilon^2}{2} - \frac{\varepsilon^3}{3})$ for $0 < \varepsilon < 1$, and a set $X$ of $p$ points in $\mathbb{R}^d$, there is a linear map $f: \mathbb{R}^d \to \mathbb{R}^k$, for all vector $\boldsymbol{u}, \boldsymbol{v} \in \boldsymbol{X}$, such that

$$(1-\varepsilon)\|\boldsymbol{u} - \boldsymbol{v}\|^2 \leq \|f(\boldsymbol{u}) - f(\boldsymbol{v})\|^2 \leq (1+\varepsilon)\|\boldsymbol{u} - \boldsymbol{v}\|^2 \qquad (1)$$

J-L lemma proves that points from high-dimensional can be embed into low-dimensional Euclidean space in a way that relative distances among the points are approximately preserved. One of the proved projection $f$ is orthogonal projection matrix proposed in [8], [9]. Firstly, a $n \times m$ random matrix is generated, and then Gram-Schmidt orthogonalization is performed to generate a Matrix $\boldsymbol{R} \in \mathbb{R}^{n \times m}$. Finally, a feature vector $\boldsymbol{x} \in \mathbb{R}^n$ is projected onto $\boldsymbol{y} \in \mathbb{R}^m$ as follows:

$$\boldsymbol{y} = \sqrt{\frac{n}{m}} \boldsymbol{R}^T \boldsymbol{x} \qquad (2)$$

RP is an effective dimension reduction method when $m < n$. In the process of Gram-Schmidt orthogonalization, the input vector needs to be linearly independent, but the generated randomly matrix does not meet this requirement. In [10], [11], the authors suggested that the RP matrix can be generated from Gaussian distributed sequences, and also proved that the generated matrix $\boldsymbol{R}$ with Gaussian distribution has the characteristic of orthogonality.

One typical application of RP is for iris [6], in this scheme, Gabor features are generated from iris image. Random projections are then applied to sectored iris feature vector, and the projected outcomes are concatenated to form a cancelable template. Even if the transformed template and the key are compromised, the original iris data cannot be retrieved thanks to the dimensionality reduction caused by Random Projection.

BioHashing [12], [13], a well-known scheme of salting based generic cancelable biometrics scheme, can be seen as an extension of random projections. In BioHashing, biometric feature $\boldsymbol{x} \in \mathbb{R}^N$ is extracted from the raw biometric data by a feature extraction method such as wavelet transform. Then $n$ orthogonal pseudo-random vectors $\boldsymbol{b}_i \in \mathbb{R}^N, i = 1, \dots, n$ ($n \leq N$) are generated with user-specific tokenized random number (TRN). The inner products are calculated between user specific pseudo-random vectors and the biometric feature vector. At last, the $n$ bit BioHash code $\boldsymbol{c}$ is computed as:

$$\boldsymbol{c} = \text{Sgn}\left(\sum \boldsymbol{x}\boldsymbol{b}_i - \tau\right) \qquad (3)$$

where $\text{Sgn}(\cdot)$ is a signum function and $\tau$ is a threshold determined empirically. The similarity between BioHash codes is calculated by Hamming distance. When a template is compromised, a new template can be generated by the same biometric feature vector and newly generated pseudo-random numbers. BioHashing can be applied on fingerprint [12], [14], iris [15], and face [13]. However, BioHashing and its variants operate under the assumption that the pseudo-random numbers would never be lost, stolen, shared or duplicated, which cannot be guaranteed in general. The performance of BioHashing will degrade dramatically under key-stolen scenario [16].

Another generic transformation based on Bloom filters is reported recently. This scheme has been applied to different popular biometric modalities including iris, face, and fingerprint [17]–[21]. The chief idea of this scheme is mapping biometric features to a bit array called Bloom filter, with several independent hashing functions. Normally a Bloom filter $b$ is a bit-array $b \in [0,1]^n$, all set to 0 initially. To embed a data set $S$ in a Bloom filter, $k$ ($k \ll n$) pre-defined independent hash functions (denoted by $h_1, h_2, \dots, h_k$) are applied to each element of S, where $k$ indices will be derived from each element. Then, all $k$ indices of bit-array $b$ are set to unity. In the case that a hash function $h(.)$ maps to index of $b$ that its value has already been set to unity, it simply ignores and proceed. Given a query element $y$, if all position $h_i(y)$ in $b$ are set as 1, $y$ can be decided to be an element of $S$, otherwise $y$ is not a member of $S$. One application of Bloom filter on iris is proposed in [21], where the binary iriscode is divided into *nBlcoks* blocks with $nBits \times nWords$ bits. A $2^{nBits}$-length Bloom filter $b$ is computed for each such block. Finally, the protected template is composed of *nBlcoks* of Bloom filters. In [20], a fingerprint template protection scheme based on Bloom filter is reported. To generate a protected template, the fingerprint original template is firstly aligned by searching a reference point, then a $N \times M$-bit binary matrix representation is generated according [22]. Finally a set of Bloom filter is built as in [21]. However, the security and privacy issue of the Bloom filter remains unresolved. For instance, a simple yet effective attack scheme that matches two template generated from the same IrisCode by different secret bit vectors, is proposed in [23] to break the criteria of non-linkability with a probability of ≥96%. In addition, a security analysis on generating false positives or recovering the key is presented: the attack complexity is $2^{25}$ for generating false positives for the smaller versions of the scheme, and a complexity between $2^2$ and $2^8$ for recovering the secret key. It is possible to fix the above issues by using non-linear and non-invertible hashing functions instead of linear mapping function, but this will degrade the efficiency of the scheme undesirably.

Index-of-Max (IoM), a newly proposed hashing method, is a ranking-based locality sensitive hashing inspired two-factor template protection technique [24]. In IoM scheme, Gaussian random matrices $W$ is firstly generated, then the product of matrices $W$ and feature vector $x$ is computed as $\bar{x}$, and finally the index of the max value in $\bar{x}$ is recorded as the hashed code. Thus, IoM transforms features from real value domain into index domain, and it strongly conceals the original biometric data. Based on the IoM concept, the authors in [24] proposed Gaussian random projection-based and uniformly random permutation-based hashing schemes, which exhibited superior performance on public benchmarks. However, IoM can only take fixed-length feature vector as input, which limits its application to variable size feature vector such as fingerprint minutiae point set.

*B. Modality-Dependent Hashing*

Since we implement our scheme and analyze fingerprint in this paper, we will only focus on fingerprint, which is also one of the most popular biometric modalities of all time.

Cappelli et al. proposed a state-of-art representation called Minutia Cylinder-Code (MCC) for fingerprint [25]. MCC is based on a 3D data structure called cylinder, which is created around each minutia point of the fingerprint. The cylinder is discretized into small cells and the contribution value of each minutia towards the cell is calculated by its position and orientation distance from the center of the cell. Since each cylinder is built based on fixed-radius cycle, it tolerates missing and spurious minutiae effectively. The use of smooth function enables it to tolerate the local distortion and small feature extraction errors. However, Ferrara et al. proposed a reconstruction strategy for MCC, and their work shows that the original MCC feature is invertible [26]. To strengthen the non-invertible property of MCC, a non-invertible MCC (P-MCC) based on dimensionality reduction and binarization is proposed in [26]. However, the revocability is not addressed in P-MCC scheme. An revocable version of P-MCC based on two-factor protection scheme, namely 2P-MCC, is proposed by Ferrara et al. [27]. In this scheme, a subset of the original bits is selected and scrambled according to a secret key, thus generating a new two-factor protected template. However, MCC is designed based on fingerprint point sets, therefore it is limited to fingerprint application, while other biometrics such as iris, face are infeasible to employ this technique.

The spectral minutiae representation proposed by Xu et al. is a fixed-length representation for fingerprint [28]–[30]. Based on the shift, scale and rotation properties of the two-dimensional continuous Fourier transform, the spectral minutiae representation is designed to be robust against translation, and rotation-scaling. In the spectral minutiae representation scheme, the fingerprint minutiae are represented as a magnitude spectrum, and it is transformed into a fixed-length feature vector, which is registration-free. However, despite the nice fixed-length characteristic, spectral minutiae representation is limited to point set based feature representation (e.g. minutiae data), and the accuracy is inferior in comparison to the state-of-art.

III. MOTIVATIONS AND CONTRIBUTIONS

As discussed above, biometric hashing techniques have some limitations, including:
- Some hashing techniques are vulnerable to certain attacks, for example, under genuine-token and stolen-token scenarios, the accuracy performance of BioHashing will deteriorate dramatically.
- Some hashing techniques are limited to certain biometric modalities. The state-of-art fingerprint minutiae protection schemes (e.g. P-MCC, 2P-MCC) are limited to point set feature data.
- The newly proposed IoM hashing scheme is limited to fixed-length feature vector, which cannot be applied to variable size feature vector.

With aforementioned discussions, we propose a generalized IoM based cancelable biometric scheme, namely gIoM which is inspired by IoM and random maxout features in machine learning study. This generalized IoM hashing transforms real-valued vector into index representation and strongly protects the original biometric data. Simultaneously, it can apply to different biometric modalities such as fingerprint, iris, and face features. In this paper, we demonstrate gIoM in fingerprint feature vector generated by a state-of-art MCC technique.

The main contributions of this paper are as follows:
1) A generalized hashing technique based on IoM is proposed and a realization on fingerprint MCC representation is implemented.
2) A full analysis of the security and privacy of gIoM hashing are presented.

IV. PRELIMINARIES

In this section we briefly introduce the random maxout features (RMF) [31] and then followed by a brief introduction of IoM, from which the proposed gIoM is extended.

*A. Random Maxout Features*

In random maxout features scheme [31], Gaussian element vectors are first generated randomly, then the input data is projected to vectors of Gaussian elements, and the maximum value of each projected sets is collected.

Let $w_j^i, i = 1 \ldots m, j = 1 \ldots q$, be independent random Gaussian vectors, drawn from $\mathbb{N}(0, I_d)$, and denote $W^i = (w_1^i \ldots w_q^i)$. For $x \in \mathbb{R}^d$, denoted $\varphi_i(x)$ as one maxout random unit:

$$\varphi_i(\mathbf{x}) = \max_{j=1 \ldots q} <w_j^i, x>, i = 1 \ldots m. \quad (4)$$

Finally, RMF vector is generated by collecting all maxout

random units:
$$\Phi(x) = \frac{1}{\sqrt{m}}[\varphi_1(x) \dots \varphi_m(x)] \in \mathbb{R}^m \quad (5)$$

### B. Index of Max Hashing

There are two realizations of IoM in [24], namely, Gaussian random based projection (GRP) and Uniform random based permutation (URP). Since our approach is mainly derived from GRP, we only introduce Gaussian random based IoM in this paper. In Gaussian random based approach, $q$ Gaussian random vectors are firstly generated $m$ times, $\{w_j^i \in \mathbb{R}^d | i = 1, \dots, m, j = 1, \dots, q\} \sim \mathbb{N}(0, I_d)$, and a random Gaussian projection matrix can be denoted as: $W^i = [w_1^i, \dots w_q^i], i = 1, \dots m$. Given a feature vector $x \in \mathbb{R}^d$, record the $m$ indices of the maximum value computed from (4) as $h_i$. Finally, the hashed code is a collection of $h_i$, and denote it as $h = \{h_i | i = 1, \dots m\}$.

## V. METHODOLOGY

We employ a well-known variable size feature vector, fingerprint Minutia Cylinder-Code (MCC) [25] as an input to demonstrate its realization. The detailed construction of MCC can be found in [25]. In this paper, MCC SDK2.0 is employed to generate the MCC templates. For each fingerprint MCC template, we extract each minutiae point's cylinder code and denote it as $c_k$. Let $C = \{c_k | k = 1, \dots N\}$ be a set of cylinder vectors of one MCC template, and let $N$ denotes the total number of minutiae points. Then the realization of gIoM is provided, followed by a matcher for the transformed vector. The process is depicted in Fig.1.

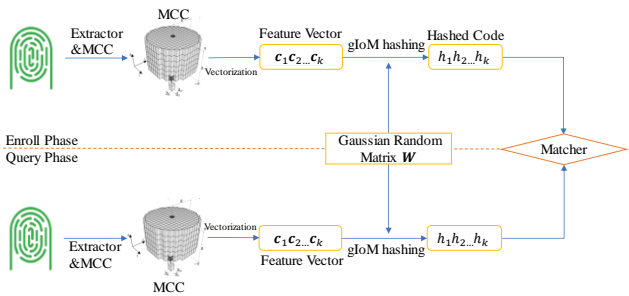

Fig.1. The process flow of the proposed gIoM Hashing.

### A. gIoM Hashing

The proposed gIoM Hashing can be summarized into two steps as follows:
1) Generate $q$ number of Gaussian random projection vectors $m$ times, denoted by $\{w_j^i \in \mathbb{R}^d | i = 1 \dots, m, j = 1, \dots, q\} \sim \mathbb{N}(0, I_d)$, and construct $m$ number of Gaussian projection matrix $W^i = [w_1^i, w_2^i \dots w_q^i]$.
2) For each cylinder vector $c_k$ in $C$, calculate the index of maximum value:
$$\hat{\varphi}_i(c_k) = \arg\max_{j=1 \dots q} < w_j^i, c_k > \quad (6)$$

Denote the index number $\hat{\varphi}_i(c_k)$ as $h_k^i$, and repeat this calculation $m$ times. The gIoM hashed code of cylinder vector set $C$ is recorded as $h = \{h_k^i \in [1, q] | i = 1, \dots m\}$.

Despite gIoM is an extended version of IoM, the iterative process presented in Step 2 of Algorithm 1 and the matching strategy differ from the original IoM in two major aspects:

1) Feature Input - gIoM accepts a variable-size feature matrix while IoM only processes the globally fixed-length feature vector.

2) Matching - Local Greedy Similarity (LGS) approach employed in gIoM while rank correlation measurement is adopted in IoM.

The pseudo-code of gIoM Hashing is given in Algorithm 1.

---
**Algorithm 1** gIoM Hashing
**Input**:
  $C$: a set of MCC cylinder vectors of one individual
  $m$: number of Gaussian random matrices
  $q$: number of Gaussian random projection vector
**Do:**
  **Step.1:** Generate $q$ number of Gaussian random projection vectors $m$ times
  **Step.2:** Perform gIoM Hashing
% For each cylinder in one MCC cylinder feature vector set
    For $k = 1: N$   % N is the total number of minutiae points
      Initialize $h_k^i$ hashed code with 0.
      For $i = 1: m$    % perform $m$ times projection
        % calculate the projected vector and
        % find the max values' index
        $\hat{\varphi}_i(c_k) = \arg\max_{j=1\dots q} < w_j^i, c_k >$
        $h_k^i = \hat{\varphi}_i(c_k)$  % $k$ is $k$th point
      End For
    End For
**Output:**
  Hashed code $h = \{h_k^i \in [1, q] | i = 1, \dots m\}$

---

### B. Matching of gIoM Hashed Codes

The gIoM hashing can preserve the relative distance in the transformed domain by converting real-value vector into indexing-based vector. To compare the similarity of two protected templates, we employ the Local Greedy Similarity (LGS) approach [32] to measure the similarity between different templates. The main idea of LGS considering two hashed code $h_A$ and $h_B$ is to compute the similarity score by averaging the matching scores of the top $n_p$ pairs with the highest scores. Specifically, the hashed code of each point is compared with another one from different sample using Euclidean distance, thus resulting in a score metric. Then, this score metric is sorted in ascending order, and the average score is computed among the top $n_p$ of the score defined in [25]. Specifically, $n_p$ is computed below according to [25]:

$$n_p = min_{n_p} + \lfloor (Z(\min\{n_A, n_B\}, \mu_P, \tau_P)) \cdot (max_{n_p} - min_{n_p}) \rfloor \quad (7)$$

where $min_{n_p}$, $max_{n_p}$, $\mu_P$ and $\tau_P$ are parameters, set as the default values as in [25]. $n_A$ and $n_B$ are the total number of minutiae points of two MCC template, and $Z$ is a sigmoid function: $Z(v, \mu, \tau) = \frac{1}{1+e^{-\tau(v-\mu)}}$.

## VI. EXPERIMENTS AND DISCUSSIONS

In this paper, MCC is adopted with the default parameters as in [25], thus the length of real value cylinder code vector is $16 \times 16 \times 6 = 1536$. MCC template mainly consists of $x, y$ coordinates (in pixels), the direction $\theta$ and cylinder cell values about each minutia. Although in the MCC scheme the $x, y$ coordinates and direction $\theta$ play important roles in the matching process, we only use the cylinder cell values since the main goal is to secure the biometric data.

Two public fingerprint databases, namely FVC2002 (DB1, DB2, DB3) and FVC2004 (DB1, DB2, DB3), are used for evaluation. For FVC2002 and FVC2004, each dataset consists of 100 users with 8 samples per user. In this paper, the accuracy performance is measured by Equal Error Rate (EER) based on the FVC2004 testing protocol:

- Each sample in the subset A is matched against the remaining samples of the same finger to compute the False Non-Match Rate (FNMR). If template $g$ is matched to $h$, the symmetric match (i.e., $h$ against $g$) is not executed to avoid correlation in the scores, and the scores is collected as genuine match score.
- The first sample of each finger in the subset A is matched against the first sample of the remaining fingers in A to compute the False Match Rate (FMR). If template $g$ is matched against $h$, the symmetric match (i.e., $h$ against $g$) is not executed to avoid correlation in the scores, and the scores is collected as imposter match score.

### A. Parameters Optimization

There are mainly two parameters in this scheme, i.e., $m$ Gaussian random matrices, and $q$ dimension of the random vector. We investigate the effect of $m$ and $q$ with respect to EER. In this experiment, $m \in \{5, 10, 50, 100, 150, 200, 250, 300, 500, 700\}$ and $q \in \{5, 10, 50, 100, 150, 200, 250, 300\}$ are considered. The average EER under different combination of $q$ and $m$ is recorded and showed in Fig. 3. Specifically, EER under different $m$ when $q = 100$ showed in Fig. 2.

Fig. 3 suggests that the EER levels off when $q$ changes from 50 to 300, and there is no significant difference for EER under different value of $m$. In other words, if we set $m$ with a large number, $q$ can be a small number without degrading the EER, which allows saving on storage and computation time.

However, $m$ is strongly correlated with EER. When $q = 100$, a better EER can be obtained when $m$ increases, and EER stabilizes when $m$ is sufficiently large. This is because a larger $m$ can produce redundant hashed codes, while a small $m$ may lead to discriminative information loss and the hash code will be dominated by random noise under this condition.

### B. Performance Evaluation

Table I tabulates the EERs of the proposed method along with the existing methods. The EERs are calculated by taking the average of EERs repeated for three times with $m = 700$, and $q = 100$. We observe that 2P-MCC (a cancelable version of MCC) outperforms others in most cases. However, 2P-MCC is specifically designed for MCC representation and cannot be applied to different biometric feature representations. On the other hand, the proposed method can, in general, achieve comparable accuracy when compared to the existing methods. Furthermore, the generalized property allows the proposed method to be utilized in any forms of biometric representation. Moreover, our method exceeds Bloom filter-based approach thanks to the distance preservation property offered by the proposed method. It is noteworthy that both [14] and [22] report the results based on a fraction of the database (e.g., only 2nd, 3rd and 6th samples in FVC2002 DB1&2 used in [14]) while the proposed method strictly follows the FVC protocol, where all 8 samples are utilized. Therefore, the accuracy performance can be justified based on the above observations.

TABLE I. Performance Accuracy (EER) and Comparison

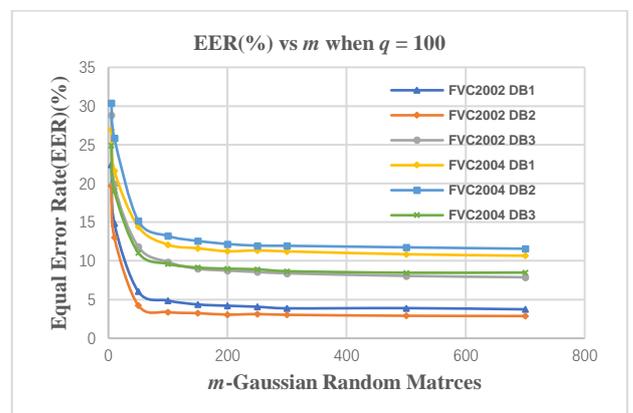

Fig. 2. The curves of EER vs $m$-Gaussian random matrices on FVC2002 (DB1-3), FVC2004 (DB1-3) when $q$=100.

| Methods | EER (%) for FVC2002 | | | EER (%) for FVC2004 | | |
|---|---|---|---|---|---|---|
| | DB1 | DB2 | DB3 | DB1 | DB2 | DB3 |
| 2P-MCC$_{64,64}$[27] | 3.3 | **1.8** | 7.8 | **6.3** | - | - |
| 2P-MCC$_{64,48}$[27] | 4.6 | 2.5 | 9.9 | 8.4 | - | - |
| Spectral Minutiae[28] | - | 3.2 | - | - | - | - |
| Bloom Filter[20][1] | 8 | 4.8 | - | - | - | - |
| Teoh et al. [12] | 15 | 15 | 27 | - | - | - |
| gIoM Hashing | 3.66 | 2.7 | **7.79** | 10.5 | 11.34 | 8.57 |

## VII. SECURITY AND PRIVACY ANALYSIS

---

[1] Average EER from different finger samples, namely, avg(3,7,14) and avg(0.5,3,11) for DB1 and DB2, respectively.

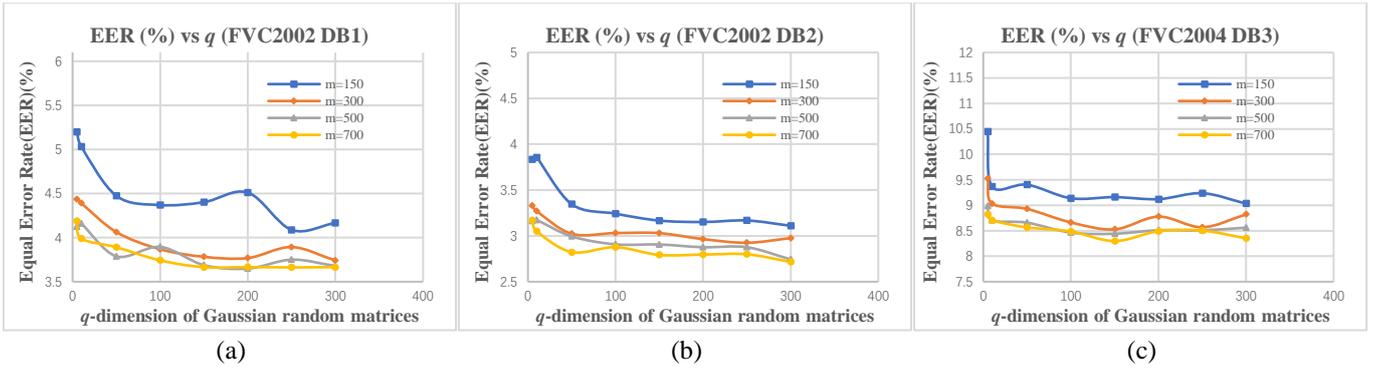

Fig.3. The curves of EER (%) vs $q$ on FVC2002 (DB1, DB2) and FVC2004 (DB3)

## A. Non-Invertibility

As stated in Section I, it should be infeasible to compute and restore the original fingerprint MCC vector from gIoM hashed code, with or without the Gaussian random matrices. Intuitively, gIoM hashed code is a collection of discrete indices, hence there is no way for an attacker to restore the original feature vector. When the attacker knows the token $W$ (Gaussian random matrix), he/she can try to retrieve the original input by solving multivariable linear inequalities. Following we provide two case studies to evaluate the non-invertibility of gIoM hashing.

*Case 1: $m$ equals to the dimension of feature vector $x$.* Let feature vector $x = [0.8\ 0.1\ 0.7]$, $m = 3$, $q = 2$, random projection matrix $W = [W_1\ W_2\ W_3]$, where

$$W_1 = \begin{pmatrix} 0.2 & -0.4 \\ -0.6 & -0.1 \\ 0.4 & 0.9 \end{pmatrix},$$
$$W_2 = \begin{pmatrix} 0.3 & 0.7 \\ 0.3 & -0.3 \\ -0.1 & 0.5 \end{pmatrix},$$
$$W_3 = \begin{pmatrix} 0.1 & -0.4 \\ 0.2 & -0.6 \\ 0.7 & 0.1 \end{pmatrix}.$$

The gIoM hashed code is generated as:
$$x \times W_1 = [0.38 \quad 0.30],\ h^1 = 1;$$
$$x \times W_2 = [0.20 \quad 0.88],\ h^2 = 2; \quad (8)$$
$$x \times W_3 = [0.59\ -0.31],\ h^3 = 1.$$

Thus, the hashed code is $h = [1\ 2\ 1]$. To restore the original vector $x = [x_1\ x_2\ x_3]$, assume the attacker knows $W$, and he/she can formulate the following inequalities:
$$0.2x_1 - 0.6x_2 + 0.4x_3 > -0.4x_1 - 0.1x_2 + 0.9x_3$$
$$0.3x_1 + 0.3x_2 - 0.1x_3 < 0.7x_1 - 0.3x_2 + 0.5x_3 \quad (9)$$
$$0.1x_1 + 0.2x_2 + 0.7x_3 > -0.4x_1 - 0.6x_2 + 0.1x_3$$
which simplifies to the following:
$$x_1 > 0, -\frac{x_1}{14} < x_2 \le \frac{14x_1}{15},$$
$$\frac{1}{3}(3x_2 - 2x_1) < x_3 < \frac{1}{5}(6x_1 - 5x_2) \quad (10)$$

With (10), although an attacker cannot determine the exact original vector, he/she can approximate the feature vector by randomly setting $x_1$ with a value and calculate other variables according to (10). The forged vector can be hashed to the same hash code since its ordinal is identical to $x$. This suggests that if an attacker can get enough inequalities he/she can plausibly break the system under the situation of case 1. More specifically, if $m$ is large enough (such as $m \ge d$, where $d$ is the dimension of $x$) it can be feasible to retrieve the ordinal information of the original vector. Thus, a vector which can be hashed to identical hash codes by gIoM, can be forged.

*Case 2: $m$ is smaller than the dimension of feature vector $x$.* Let the feature vector $= [0.8\ 0.1\ 0.7\ 0.5]$, $m = 2$, $q = 2$, random projection matrix $W = [W_1\ W_2]$,

$$W_1 = \begin{pmatrix} 0.2 & -0.4 \\ -0.6 & -0.1 \\ 0.4 & 0.9 \\ 0.9 & 0.5 \end{pmatrix},$$
$$W_2 = \begin{pmatrix} 0.3 & 0.7 \\ 0.3 & -0.3 \\ -0.1 & 0.5 \\ 0.4 & 0.1 \end{pmatrix}.$$

The gIoM hashed code is $h = [1\ 2]$. Let $x = [x_1\ x_2\ x_3\ x_4]$, and formulate the following inequalities:
$$0.2x_1 - 0.6x_2 + 0.4x_3 + 0.9x_4 >$$
$$-0.4x_1 - 0.1x_2 + 0.9x_3 + 0.5x_4$$
$$0.3x_1 + 0.3x_2 - 0.1x_3 + 0.4x_4 < \quad (11)$$
$$0.7x_1 - 0.3x_2 + 0.5x_3 + 0.1x_4$$
which simplifies to the following:
$$x_2 < \frac{9x_4 + 56x_1}{60},$$
$$\frac{3x_4 - 4x_1 + 6x_2}{6} < x_3 < \frac{4x_4 + 6x_1 - 5x_2}{5} \quad (12)$$

Under this case, even the attacker can learn the relationship among $x_1\ x_2\ x_3\ x_4$ from (12), it is challenging to decide the direct relationship among each variable. Brute force attack is still needed to guess the original vector and the relationship. If attacker decides to guess the real-value directly, we assume the attacker knows well about the input vector. Note that the minimum and maximum values of the cylinder cell value are 0.000 and 1.000, respectively according to the MCC algorithm. For one value of the feature vector, it has $10^4$ possibilities, since the length of real value MCC cylinder code is of 1536 dimensions. Consequently, the total possibilities to guess the entire original feature vector requires $10^{4 \times 1536}$ attempts. Therefore, even with (12), it is still infeasible to have a clue in guessing the original real number. Indeed, when $m \ll d$, gIoM is an effective dimension reduction method. Therefore, it can strongly conceal the original biometric information.

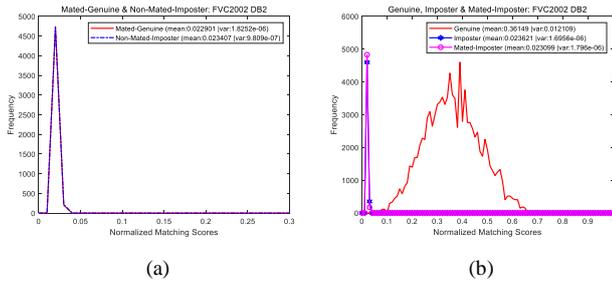

Fig.4. (a) Distributions of Mated-Genuine and Non-Mated-Imposter for non-linkability analysis. It is infeasible to differentiate two hashed codes since the score distribution is overlapping. (b) Distributions of Non-Mated-Imposter, Genuine and Imposter for revocability analysis. It is infeasible to attack a renewed template with any revoked templates since distribution of Non-Mated-Imposter is far from Genuine scores.

To sum up, the non-invertibility of gIoM can be ensured when $m$ is largely smaller than the dimension of the original feature vector, leading to an unsolvable system of inequalities as illustrated above. Since the dimension of MCC vector is 1536 while $m$ is at most 700, the non-invertibility of gIoM is ensured.

*B. Non-Linkability*

To validate the requirement of non-linkability, we first introduce the Mated-Genuine scores and Non-Mated-Imposter scores in this section. The Mated-Genuine score is the matching score between two gIoM hashed codes generated from the same fingerprint by employing two different Gaussian random matrices, while the Non-Mated-Imposter score is generated by matching two gIoM hashed codes generated from two different fingerprints by employing two different Gaussian random matrices.

In this context, we assume that the attacker: (a) knows well about gIoM scheme; (b) is in possession of two gIoM hashed codes from different databases or application, and; (c) can calculate the matching score between two gIoM hashed codes. Under this scenario, if the Non-Mated-Imposter and Mated-Genuine score distributions are different, then the attacker can easily decide whether two hashed codes are from same individual. To prove the non-linkability of gIoM hashing, FVC2002 (DB2) is used to test the score distributions. From Fig. 4(a), it is clear that Non-Mated-Imposter and Mated-Genuine score distributions are overlapped largely, which implies that even an attacker can get all matching scores, he/she is still unable to decide whether two hashed codes are the same individual, thus proving the criteria of non-linkability for gIoM hashing.

*C. Revocability*

Once the biometric template in the database is compromised, a new protected template should be generated. In the proposed gIoM hashing scheme, a new hashed code can be easily generated with a new Gaussian random matrix. In real life scenario, the Gaussian random matrices are usually user-specific for revocability. Therefore, when one user's biometric template is stolen, only this user's Gaussian random matrices needs to be regenerated. To demonstrate the security of renewed template, the distributions of Mated-Genuine, Genuine and Imposter scores have been evaluated on FVC2002 (DB2). To generate Mated-Genuine score, 50 sets of Gaussian random matrices are generated for each user's first fingerprint image. Subsequently, 50 Mated-Genuine hashing codes are generated using the first fingerprint of each user with 50 different Gaussian random matrices. Those 50 hashing codes are then matched against the original hashing codes of this fingerprint to produce 50 Mated-Genuine scores. Altogether 100×50=5000 Mated-Genuine scores are collected for all 100 users. As observed in Fig.4 (b), the distribution of Mated-Genuine is largely overlapped with the distribution of imposter scores, which means that there is no difference between templates generated by different random matrices on the same individual biometric or different individual biometric.

## VIII. CONCLUSIONS

A generic gIoM hashing algorithm is proposed in this paper. We demonstrate a realization of gIoM hashing for fingerprint MCC vectors. The accuracy performance of gIoM hashing is shown to be preserved theoretically and empirically. gIoM hashing is also designed to meet the template protection criteria, i.e., non-linkability, revocability and non-invertibility, which have been analyzed in this paper.

Our future work will focus on the output of hashed code, where a robust algorithm will be devised to yield fixed-length hashed output even the dimension of the input vector is variable. Secondly, as a cancelable biometric scheme, gIoM hashing can be applied to verification (1 to many) scenario, but the application to identification (many to many) still needs further investigations.